\documentclass[12pt]{gsis_arxiv}  
\usepackage[misc]{ifsym} 
\usepackage{graphicx,subfigure}
\usepackage{hyperref,caption}
\usepackage[para,online,flushleft]{threeparttable}
\usepackage{multirow,booktabs}
\usepackage{amsmath,amssymb}
\usepackage{geometry}
\usepackage{makecell}
\usepackage[table]{xcolor}
\definecolor{bestgreen}{RGB}{146,208,80}
\definecolor{secondgreen}{RGB}{198,239,206}
\definecolor{thirdyellow}{RGB}{255,242,204}
\usepackage{natbib}
\bibpunct[, ]{(}{)}{;}{a}{}{,}
\begin{document}
\pagestyle{fancy}
\fancyhf{}
\fancyfoot[C]{\thepage}
\renewcommand{\headrulewidth}{0pt}

\title{DECO: Depth-Guided Co-Visibility Reasoning for Low-Altitude UAV Visual Localization
}

\author{Yibin Ye}
\author{Xichao Teng}
\author{Shuo Chen}
\author{Xiaokai Song}
\author{Dongdong Guan}
\author{Qifeng Yu}
\author{Zhang Li \Letter}

\affil{College of Aerospace Science and Engineering, National University of Defense Technology, Changsha 410073, China}

\thanks{\textbf{CONTACT:} Zhang Li \quad \Letter : zhangli\_nudt@163.com}


\Abstract{
Unmanned aerial vehicles (UAVs) increasingly require robust visual localization in GNSS-denied environments. A common solution estimates UAV poses by matching keypoints between UAV images and geo-tagged orthographic reference maps derived from satellite or aerial imagery, followed by Perspective-\(n\)-Point (PnP) pose solving. However, such reference maps mainly record top-down surfaces such as roofs and ground planes, while vertical structures such as facades and walls are often compressed or missing. Consequently, many visually distinctive keypoints in low-altitude UAV images have no valid counterparts in the reference map, leading to redundant matches and inaccurate pose estimation. To address this issue, we propose DECO, a \textbf{DE}pth-guided \textbf{CO}-visibility reasoning framework for low-altitude UAV visual localization. DECO uses monocular depth priors to infer local surface geometry and estimate co-visible regions between UAV images and the reference map. Based on this prior, a Geometry-Saliency Coupled Co-visibility Score is introduced to jointly consider geometric co-visibility and detector saliency for keypoint ranking. In this way, DECO retains keypoints that are both visually distinctive and geometrically co-visible, improving feature matching and PnP-based pose estimation. Extensive experiments demonstrate that DECO achieves superior localization performance and can be integrated with different depth models, feature detectors, and matchers. The source code will be available at \url{https://github.com/UAV-AVL/DECO}.

}

\maketitle



\section{Introduction}

Unmanned aerial vehicles (UAVs) are increasingly used in remote sensing applications such as disaster response, urban mapping, and infrastructure inspection, where reliable georeferencing is essential for safe and accurate mission execution \citep{li2018uav, chang2023review}. Current UAV georeferencing systems mainly rely on global navigation satellite systems (GNSS) and inertial navigation systems (INS) \citep{GNSS_1, GNSS_INS}. However, GNSS signals can be degraded by terrain occlusion and electromagnetic interference, while INS inevitably accumulates drift over time \citep{yin2024sky, he2024leveraging}. These limitations have motivated increasing interest in absolute visual localization for GNSS-denied environments \citep{lu2018survey}. A common solution is to match UAV images with geo-tagged orthographic reference maps, such as orthorectified satellite imagery or aerial orthophotos \citep{chen2021real, ye2024coarse}. When a co-registered digital surface model (DSM) is available, matched points in the reference map can be lifted to 3D world points, thereby converting 2D--2D matches into 2D--3D correspondences. The UAV pose is then estimated by solving a Perspective-\(n\)-Point (PnP) problem \citep{ye2025exploring, dhaouadi2025ortholoc}.

Although absolute visual localization has been widely studied, many existing methods are designed for high-altitude, near-nadir imaging conditions \citep{chang2023review}. Under these conditions, UAV images and orthographic reference maps usually observe the scene from similar top-down viewpoints, so their geometric relationship can often be approximated by a near-planar transformation \citep{MI}. However, with the rapid growth of low-altitude UAV applications \citep{wang2025toward}, UAVs are increasingly required to operate below 300~m \citep{ye2025exploring} and capture oblique images for tasks such as infrastructure inspection and target reconnaissance. Under low-altitude oblique imaging conditions, UAV images often contain many vertical structures, such as building facades, walls, and object sides. Geo-tagged 2D reference maps, however, retain only limited evidence of these vertical structures because they are formed through a top-down orthographic projection. This difference leads to a co-visibility gap between the UAV image and the reference map. Regions that are highly salient in the UAV image may therefore be invalid for UAV-to-map matching because they are not observable from the top-down reference view. These regions can expand the matching search space, introduce unreliable correspondences, and weaken the geometric consistency required for PnP-based pose estimation.

To improve pose estimation accuracy, considerable effort has been devoted to robust cross-view image matching. Among existing solutions, detector-based matching remains an important technical route because of its flexibility and interpretability \citep{SIFT,detone2018superpoint,zhao2023aliked}. In this pipeline, keypoints are first detected in both images, represented by local descriptors, and then matched according to descriptor similarity and geometric consistency. Classical handcrafted methods, such as SIFT, SURF, and affine-invariant extensions \citep{SIFT, surf, asift}, often degrade significantly under large viewpoint changes, limiting their effectiveness for low-altitude UAV visual localization. More recently, deep learning has substantially improved the robustness of cross-view matching, with many studies focusing on stronger feature detectors \citep{potje2024xfeat, alike} and more reliable feature matchers \citep{superglue, lightglue, omniglue}. Nevertheless, most of these methods focus on improving local feature detection, description, or matching, without explicitly reasoning about region-level co-visibility between the UAV image and the orthographic reference map before matching. As a result, visually salient but geometrically non-co-visible keypoints may still be selected and matched, weakening the reliability of the final pose estimation.

Some recent studies have attempted to suppress invalid matching regions before point-level matching by estimating image overlap \citep{chen2022guide, pan2025scale} or relative scale \citep{barroso2022scalenet}, while others use semantic cues to identify semantically consistent regions \citep{SGAM, MESA, dmesa}. However, these strategies are mainly designed for ground-level or generic cross-view image pairs. In low-altitude UAV visual localization, the large viewpoint difference between UAV images and reference maps, together with the complexity of remote sensing scenes, makes it difficult to reliably estimate overlapping or semantically consistent regions. In addition to semantic priors, geometric priors derived from monocular depth estimation (MDE) have also been introduced into image matching. This direction has recently become more practical with the emergence of zero-shot MDE models, which can provide robust relative geometry and surface-orientation cues without task-specific training. For example, some methods divide the scene into multiple local planes for separate matching \citep{toft2020single, huang2023deep}, while others incorporate depth information into feature extraction and matching \citep{wang2023guiding, LiftFeat}. However, these methods mainly use depth to improve matching robustness under viewpoint changes, rather than to explicitly reason about the co-visibility between the UAV image and the orthographic reference map. Therefore, the matching ambiguity caused by the co-visibility gap remains largely unresolved.

In this paper, we propose DECO, a \textbf{DE}pth-guided \textbf{CO}-visibility reasoning framework for low-altitude UAV visual localization. Instead of directly matching all detected keypoints, DECO first estimates whether local regions in the UAV image are likely to be co-visible with the top-down reference map. Specifically, DECO uses monocular depth priors to recover local surface geometry and derive a geometric co-visibility prior between the UAV image and the orthographic reference map. Based on this prior, we introduce a Geometry-Saliency Coupled Co-visibility Score (GS-CoVis score) for keypoint ranking. This score combines geometric co-visibility with detector saliency, allowing DECO to retain keypoints that are both visually distinctive and geometrically consistent with the reference map. By placing co-visibility reasoning before feature matching, DECO reduces redundant correspondences and improves the reliability of subsequent PnP-based pose estimation. The framework is training-free and can be integrated with different MDE models, feature detectors, and feature matchers.

The main contributions of this work are summarized as follows:
\begin{itemize}
    \item We identify the co-visibility gap as a key challenge in low-altitude UAV visual localization with orthographic reference maps. To address this challenge, we propose DECO, a depth-guided co-visibility reasoning framework that estimates geometric consistency between the UAV image and the top-down reference view before feature matching.

    \item We introduce the GS-CoVis score for keypoint ranking. The score jointly considers geometric co-visibility and detector saliency, allowing DECO to retain keypoints that are both visually distinctive and geometrically consistent for UAV-to-map matching.

    \item Extensive experiments demonstrate that DECO achieves superior localization performance compared with state-of-the-art methods. Further evaluations with different MDE models, feature detectors, and feature matchers show that DECO can serve as a plug-and-play module for improving low-altitude UAV visual localization.
\end{itemize}

\section{Related Work}
\subsection{Cross-View Image Matching}

Cross-view image matching has evolved from handcrafted pipelines to deep-learning-based approaches \citep{he2025matchanything}. Existing deep-learning-based methods can be broadly divided into detector-based and detector-free pipelines. Detector-based methods establish sparse correspondences by detecting and describing keypoints in each image. Representative advances include SuperPoint, DISK, ALIKED, and XFeat for feature detection and description, as well as SuperGlue and LightGlue for feature matching \citep{detone2018superpoint,DISK,zhao2023aliked,potje2024xfeat,superglue,lightglue}. By contrast, detector-free methods avoid explicit keypoint detection and directly predict dense or semi-dense correspondences in an end-to-end manner. Representative models include LoFTR, Efficient LoFTR, DKM, and RoMa \citep{sun2021loftr,wang2024efficient,DKM,edstedt2024roma}. Detector-free methods often show stronger robustness in weak-texture regions and under severe viewpoint changes, but usually require higher computational and memory costs \citep{ye2025exploring}. For onboard UAV localization, detector-based pipelines therefore remain attractive because they are generally more efficient and easier to deploy in practice \citep{chen2021real, ye2024coarse}. However, in low-altitude UAV visual localization, detector-based methods have an inherent limitation: since keypoints are detected independently in each image, the detector cannot determine in advance which regions are truly co-visible between the UAV image and the orthographic reference map without additional prior information \citep{chen2022guide, dmesa}. As a result, many detected keypoints may fall on vertical structures or other regions that have no valid counterparts in the orthographic reference map, introducing redundant correspondences and degrading downstream pose estimation accuracy.

\subsection{Overlap and Semantic Priors for Image Matching}

To alleviate this limitation, recent studies have introduced additional priors before point-level correspondence estimation. One line of work identifies co-visible or geometrically consistent regions before feature matching. Typical strategies include overlap estimation to suppress invalid matching regions in advance \citep{chen2022guide, pan2025scale}, relative-scale estimation to reduce scale ambiguity \citep{barroso2022scalenet}, and co-visibility-aware modeling during correspondence construction, as in CoMatch \citep{li2025comatch}. By constraining feature extraction or correspondence search to regions that are more likely to be jointly observed, these methods improve matching robustness under challenging viewpoint and scale changes. Another line of work introduces semantic priors for area-to-point matching, where semantically consistent regions are first localized and then used to guide point-level correspondence estimation, as represented by SGAM, MESA, and DMESA \citep{SGAM, MESA, dmesa}. Although these methods have shown promising results on ground-level benchmarks \citep{MegaDepthLi18}, their underlying assumptions become harder to satisfy in low-altitude UAV visual localization. In such scenarios, UAV images and orthographic reference maps often exhibit large viewpoint differences, strong perspective distortion, and complex scene content, making reliable overlap estimation and stable semantic correspondence substantially more difficult across views \citep{liu2025diffusionuavloc, zeng2025segmatch}.

\subsection{Monocular-Depth Priors for Image Matching}

Besides overlap and semantic priors, geometric priors have also been introduced to mitigate cross-view geometric discrepancies. In particular, recent advances in monocular depth estimation models have shown strong spatial understanding and zero-shot generalization capabilities \citep{DA3, depthpro, piccinelli2025unidepthv2}, motivating their use as geometric priors for image matching. Existing studies exploit monocular depth priors in different ways. Toft \emph{et al.} \citep{toft2020single} used monocular depth predictions to identify planar regions and rectify perspective distortion before feature extraction, while Huang \emph{et al.} \citep{huang2023deep} introduced depth-aware homography estimation for image matching in non-planar scenes. Wang \emph{et al.} \citep{wang2023guiding} used surface curvature estimated from predicted depth as an additional cue for feature matching; Karpur \emph{et al.} \citep{karpur2024lfm} incorporated 3D signals, including monocular depth, into a learnable matcher for wide-baseline correspondence estimation; and Lee \emph{et al.} \citep{lee2025depth} introduced depth-guided pseudo-3D constraints for local outlier rejection in urban UAV image matching. Monocular depth has also been used to enhance local feature representation; for example, LiftFeat \citep{LiftFeat} uses depth-derived surface normals to learn 3D geometry-aware descriptors and improve feature distinctiveness in challenging conditions. However, existing depth-guided methods mainly use monocular depth to improve feature representation, geometric transformation estimation, or correspondence filtering. In contrast, DECO exploits depth-derived surface geometry for explicit cross-view co-visibility reasoning and co-visibility-aware keypoint ranking in UAV-to-map localization.


\section{Methodology}
\label{sec:method}

\subsection{Problem Formulation and Framework Overview}

\begin{figure}[htb]
    \centering
    \includegraphics[width=\textwidth]{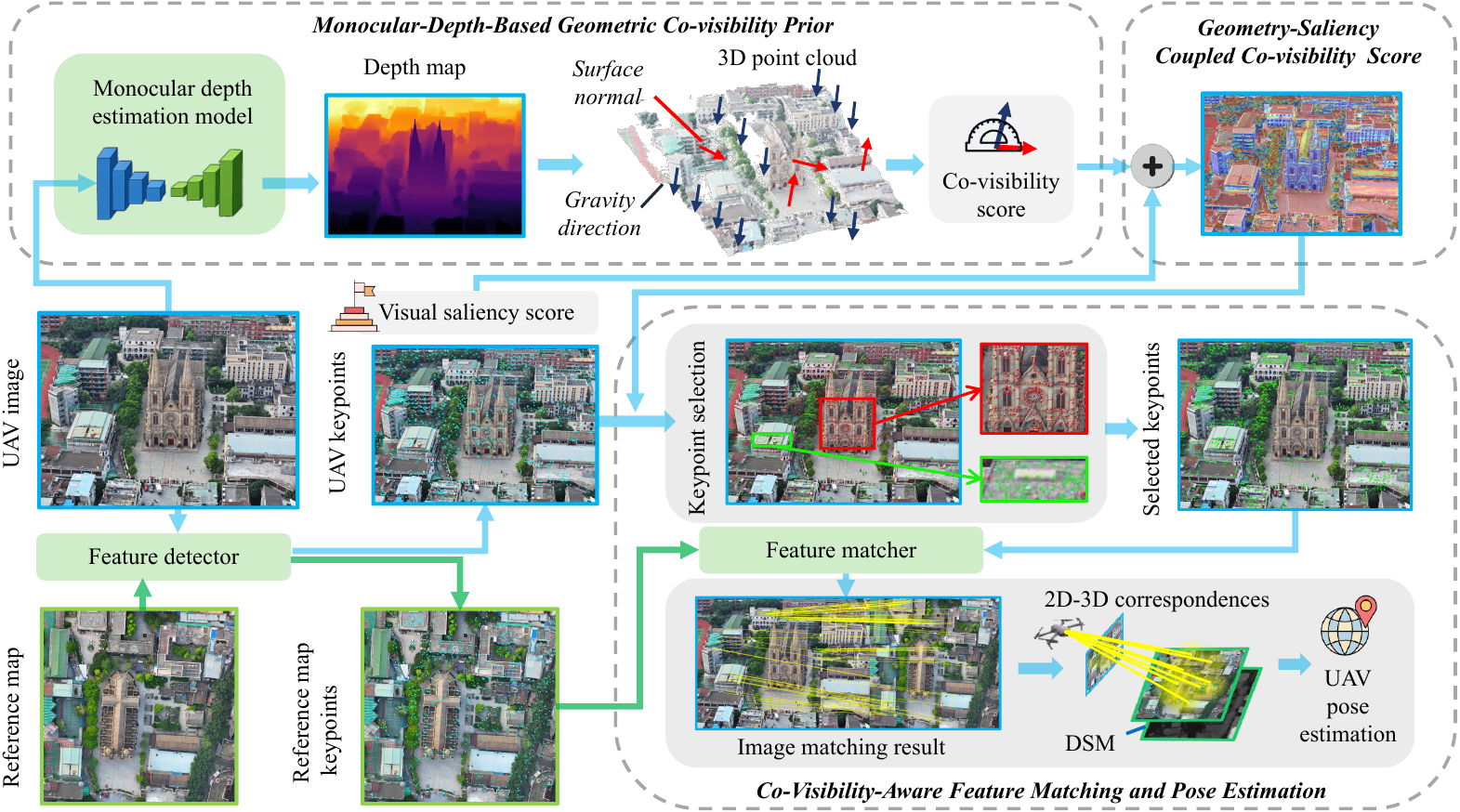}
    \caption{blue}{Overview of the proposed DECO framework.}

    \label{fig:framework}
\end{figure}

As illustrated in Figure~\ref{fig:framework}, the DECO framework consists of three main steps. First, DECO predicts monocular depth from the UAV image and derives a dense geometric co-visibility prior by comparing local surface normals with the gravity direction. Second, this geometric prior is coupled with detector saliency to compute the GS-CoVis score, which ranks UAV keypoints by considering both geometric co-visibility and visual distinctiveness. Finally, high-scoring UAV keypoints are selected and matched with the reference map, and the matched reference-map points are lifted to 3D using the DSM for pose estimation. 

DECO operates on three primary data sources: a UAV image $I \in \mathbb{R}^{H \times W}$, a georeferenced reference map $R \in \mathbb{R}^{H_r \times W_r}$, and a co-registered DSM. In addition, the camera intrinsic matrix $\mathbf{K}$ and a coarse attitude prior are assumed to be available as known auxiliary parameters. The camera intrinsic matrix is
\begin{equation}
\mathbf{K}=
\begin{bmatrix}
f_x & 0 & c_x\\
0 & f_y & c_y\\
0 & 0 & 1
\end{bmatrix},
\label{eq:K}
\end{equation}
where $f_x$ and $f_y$ are the focal lengths and $(c_x,c_y)$ is the principal point. The coarse attitude prior consists of the roll angle $\phi$ and the pitch angle $\theta$, which are commonly available from the onboard inertial measurement unit (IMU). In DECO, this prior is used only to estimate the gravity direction in the camera frame, rather than to directly constrain the final pose. The pitch angle is defined relative to the horizontal plane, with downward tilt being positive.
\subsection{Monocular-Depth-Based Geometric Co-visibility Prior}

An MDE model is first used to construct a dense geometric prior. This prior measures whether local regions in the UAV image are likely to be visible in the orthographic reference map. In the main implementation, we adopt Depth Anything 3 (DA3)~\citep{DA3} as the default depth estimator, while the generality of DECO with other MDE models is evaluated in Section~\ref{sec:generality}.

Let $\mathcal{F}_{\mathrm{mde}}(\cdot)$ denote the MDE model. Given the UAV image $I$, the predicted depth map is $Z=\mathcal{F}_{\mathrm{mde}}(I)$, where $Z(\mathbf{p})$ is the estimated depth at pixel $\mathbf{p}=(u,v)^\top \in \Omega_I$. Let $\mathbf{X}_c(\mathbf{p}) \in \mathbb{R}^3$ denote the 3D point corresponding to pixel $\mathbf{p}$ in the camera coordinate system. For each pixel with homogeneous coordinate $\tilde{\mathbf{p}}=[u,v,1]^\top$, $\mathbf{X}_c(\mathbf{p})$ is obtained by back-projection:

\begin{equation}
\mathbf{X}_c(\mathbf{p})
=
Z(\mathbf{p})\mathbf{K}^{-1}\tilde{\mathbf{p}}
=
\left[
\frac{(u-c_x)Z(\mathbf{p})}{f_x},
\frac{(v-c_y)Z(\mathbf{p})}{f_y},
Z(\mathbf{p})
\right]^{\top}.
\label{eq:backprojection}
\end{equation}
The reconstructed local point cloud is $\mathcal{P}_c=\{\mathbf{X}_c(\mathbf{p}) \mid \mathbf{p}\in\Omega_I\}$.

A surface-normal field is then estimated from $\mathcal{P}_c$ in the camera coordinate system. For each valid pixel $\mathbf{p}=(u,v)$ whose neighboring depth values are available, the local tangent directions are approximated by finite differences:
\begin{equation}
\mathbf{t}_u(\mathbf{p}) =
\mathbf{X}_c(u+\delta,v)-\mathbf{X}_c(u-\delta,v),
\qquad
\mathbf{t}_v(\mathbf{p}) =
\mathbf{X}_c(u,v+\delta)-\mathbf{X}_c(u,v-\delta),
\label{eq:tangent}
\end{equation}
where $\delta$ is the finite-difference step size. The local surface normal is computed as
\begin{equation}
\mathbf{n}(\mathbf{p}) =
\frac{\mathbf{t}_u(\mathbf{p}) \times \mathbf{t}_v(\mathbf{p})}
{\left\| \mathbf{t}_u(\mathbf{p}) \times \mathbf{t}_v(\mathbf{p}) \right\|_2}.
\label{eq:normal}
\end{equation}
Pixels with invalid depth values are excluded.

Each local surface normal is then compared with the gravity direction. Let the gravity direction in the world coordinate system be $\mathbf{g}_w=[0,0,1]^\top$. Following the roll--pitch convention in the attitude metadata, the world-to-camera tilt rotation is approximated as $\mathbf{R}_{wc}=\mathbf{R}_x(\phi)\mathbf{R}_y(\theta)$, where $\mathbf{R}_x(\phi)$ and $\mathbf{R}_y(\theta)$ denote rotations induced by roll and pitch. The camera-frame gravity direction is 
\begin{equation}
\mathbf{g}_c=\mathbf{R}_{wc}\mathbf{g}_w.
\label{eq:gravity}
\end{equation}
The dense geometric co-visibility score is defined as the absolute cosine similarity between the local surface normal and the camera-frame gravity direction:
\begin{equation}
a(\mathbf{p}) =
\left|
\frac{\mathbf{n}(\mathbf{p})^\top \mathbf{g}_c}
{\|\mathbf{n}(\mathbf{p})\|_2 \|\mathbf{g}_c\|_2}
\right|,
\label{eq:normal_gravity_score}
\end{equation}
where $a(\mathbf{p})\in[0,1]$. A larger value indicates that the local surface is more likely to be co-visible with the orthographic reference map.

To improve spatial consistency and suppress isolated noisy responses, morphological opening followed by closing is applied directly to the continuous dense score map, without thresholding or binarization. Specifically, both operations use a $5\times5$ elliptical structuring element and are applied once:
\begin{equation}
\hat{a} =
\mathcal{C}\left(
\mathcal{O}\left(a\right)
\right),
\end{equation}
where $\mathcal{O}(\cdot)$ and $\mathcal{C}(\cdot)$ denote morphological opening and closing, respectively. The $5\times5$ kernel size was empirically selected and kept fixed across all experiments.


\subsection{Geometry-Saliency Coupled Co-visibility Score}

The dense prior $\hat{a}$ describes region-level geometric co-visibility, whereas feature matching is performed at sparse keypoints. This prior is therefore sampled at keypoint locations and coupled with detector saliency. The two cues play complementary roles. Detector confidence favors visually distinctive points, but some of these points may lie on facades or other regions that are weakly represented in the orthographic reference map. In contrast, geometric co-visibility favors regions with better cross-view consistency, but these regions may not always be distinctive enough for descriptor matching.

Let $\mathcal{E}(\cdot)$ denote the feature detector. Applying $\mathcal{E}$ to the UAV image yields
\begin{equation}
\mathcal{S}_I =
\mathcal{E}(I)
=
\left\{
(\mathbf{p}_i,\mathbf{d}_i,s_i)
\right\}_{i=1}^{N},
\label{eq:uav_features}
\end{equation}
where $\mathbf{p}_i$ is the keypoint location, $\mathbf{d}_i$ is the descriptor, and $s_i$ is the detector confidence score. For each UAV keypoint $\mathbf{p}_i$, the keypoint-level geometric co-visibility score is sampled from the refined prior:
\begin{equation}
a_i = \hat{a}(\mathbf{p}_i).
\label{eq:keypoint_cov_score}
\end{equation}

Different detectors may produce confidence scores with different ranges and distributions. To reduce this dependence, the raw detector confidence scores are converted into rank-normalized saliency scores:
\begin{equation}
\rho_i =
1 -
\frac{\operatorname{rank}_{\downarrow}(s_i)}
{N-1},
\label{eq:rank_normalization}
\end{equation}
where $\operatorname{rank}_{\downarrow}(s_i)$ is the descending rank of $s_i$ among all UAV keypoints, with the most confident keypoint assigned rank $0$. This rank-based normalization makes the saliency term less dependent on the score scale of a specific detector.

The Geometry-Saliency Coupled Co-visibility Score (GS-CoVis score) is defined as
\begin{equation}
w_i =
a_i^{\lambda}\rho_i,
\label{eq:weighted_keypoint_score}
\end{equation}
where $\lambda$ controls the relative emphasis between geometric co-visibility and detector saliency. When $\lambda<1$, the ranking becomes closer to saliency-dominated selection, whereas when $\lambda>1$, more emphasis is placed on the geometric prior. A high value of $w_i$ indicates that the keypoint is both geometrically co-visible and visually distinctive. The influence of $\lambda$ on the GS-CoVis score is further discussed in Section~\ref{sec:sensiti}.

DECO ranks UAV keypoints according to $w_i$ and retains the top-$K$ keypoints:
\begin{equation}
\mathcal{S}_I^{K}
=
\operatorname{TopK}
\left(
\mathcal{S}_I, w_i, K
\right).
\label{eq:topk_weighted_selection}
\end{equation}
Thus, the dense prior $\hat{a}$, the keypoint-level score $a_i$, and the final coupled score $w_i$ define a progressive ranking process from image regions to sparse keypoints.

\subsection{Co-Visibility-Aware Feature Matching and Pose Estimation}

After keypoint ranking, the selected UAV keypoints are matched with reference-map keypoints. Applying the same feature detector to the reference map yields
\begin{equation}
\mathcal{S}_R =
\mathcal{E}(R)
=
\left\{
(\mathbf{q}_j,\mathbf{e}_j,r_j)
\right\}_{j=1}^{M},
\label{eq:ref_features}
\end{equation}
where $\mathbf{q}_j$, $\mathbf{e}_j$, and $r_j$ denote the reference-map keypoint location, descriptor, and detector confidence score.

The selected UAV keypoints are matched with reference-map keypoints using a feature matcher $\mathcal{M}_{\mathrm{feat}}(\cdot)$:
\begin{equation}
\mathcal{M}_{2D}
=
\mathcal{M}_{\mathrm{feat}}
\left(
\mathcal{S}_I^{K},
\mathcal{S}_R
\right)
=
\left\{
(\mathbf{p}_k,\mathbf{q}_k)
\right\}_{k=1}^{N_m},
\label{eq:feature_matching}
\end{equation}
where $\mathcal{M}_{2D}$ denotes the resulting 2D--2D correspondences. Since UAV keypoints have been ranked and filtered by the GS-CoVis score, matching is focused on keypoints that are more likely to have valid counterparts in the reference map.

For pose estimation, the DSM co-registered with the reference map is used to lift matched reference-map keypoints to 3D. Let $D_R:\Omega_R\rightarrow\mathbb{R}$ denote the DSM, where $D_R(\mathbf{q})$ gives the elevation at reference-map pixel $\mathbf{q}$. Let $\mathcal{G}:\Omega_R\rightarrow\mathbb{R}^2$ denote the georeferencing function that maps a reference-map pixel to its planar world coordinate. For each matched reference-map keypoint $\mathbf{q}_k$, the corresponding 3D world point is
\begin{equation}
\mathbf{X}_k^{w} =
\begin{bmatrix}
\mathcal{G}(\mathbf{q}_k) \\
D_R(\mathbf{q}_k)
\end{bmatrix}.
\label{eq:world_point}
\end{equation}
Together with the matched UAV keypoint $\mathbf{p}_k$, this forms the 2D--3D correspondence set
\begin{equation}
\mathcal{C}
=
\left\{
\left(\mathbf{p}_k,\mathbf{X}_k^{w}\right)
\right\}_{k=1}^{N_m}.
\label{eq:correspondence_set}
\end{equation}

Given $\mathcal{C}$ and $\mathbf{K}$, the UAV pose is estimated by PnP with RANSAC. Let $\mathbf{R}$ and $\mathbf{t}$ denote the world-to-camera rotation and translation. The pose is obtained by minimizing the reprojection error:
\begin{equation}
(\mathbf{R}^{\ast},\mathbf{t}^{\ast})
=
\arg\min_{\mathbf{R},\mathbf{t}}
\sum_{k=1}^{N_m}
\left\|
\mathbf{p}_k -
\pi\!\left(\mathbf{K},\mathbf{R},\mathbf{t},\mathbf{X}_k^{w}\right)
\right\|_2^2,
\label{eq:reprojection}
\end{equation}
where $\pi(\cdot)$ denotes the camera projection function. The camera center in the world coordinate system is recovered as $\mathbf{c}^{\ast}=-(\mathbf{R}^{\ast})^{\top}\mathbf{t}^{\ast}$. The planar components of $\mathbf{c}^{\ast}$ are then converted into geographic coordinates to obtain the final UAV location.

\section{UAV Visual Localization Experiments}
\subsection{Experimental Setup}
\subsubsection{Datasets}

To comprehensively evaluate the proposed method under low-altitude imaging conditions, we conduct UAV visual localization experiments on two benchmarks, namely AnyVisLoc~\citep{ye2025exploring} and OrthoLoC~\citep{dhaouadi2025ortholoc}. Both datasets contain low-altitude UAV images captured in diverse scenes, together with corresponding 2.5D reference data, including reference maps and DSMs. AnyVisLoc contains 18,000 samples collected from 25 locations across 15 cities in China, while OrthoLoC consists of 16,427 samples from 47 locations across 19 regions in Germany and the United States.  

\begin{figure}[htb]
    \centering
    \includegraphics[width=\textwidth]{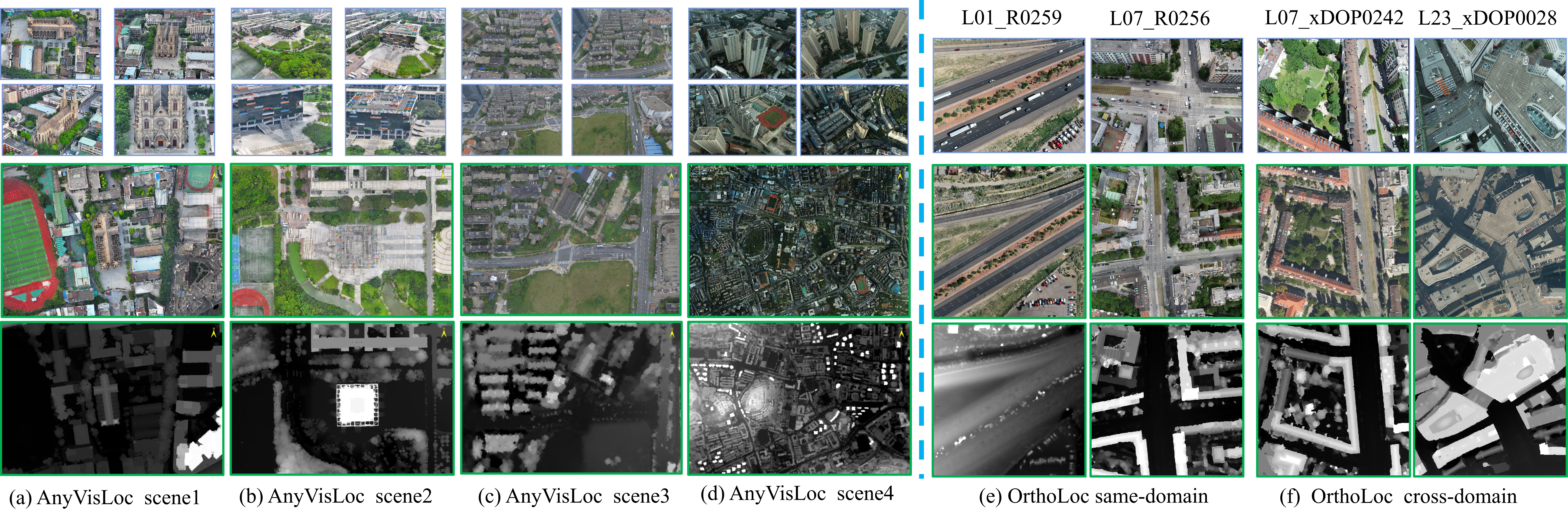}
    \caption{Example images from the test datasets. (a)–(d) are sampled from the AnyVisLoc dataset, and (e)–(f) from the OrthoLoC dataset.}
    \label{fig:dataset_overview}
\end{figure}

To ensure a focused and fair evaluation of the proposed DECO framework, we construct the test data according to the evaluation protocol and the characteristics of each dataset. Table~\ref{tab:dataset_info} summarizes the basic characteristics of the test data. For AnyVisLoc, we use a representative subset consisting of four typical urban scenes, with 2,492 image pairs in total. Examples of these four scenes are shown in Figure~\ref{fig:dataset_overview}(a)--(d). These scenes cover diverse camera settings, flight altitudes, pitch angles, and urban structures. In particular, most UAV images have pitch angles between $0^\circ$ and $60^\circ$. These oblique views contain pronounced elevation variations and many vertical structures, such as building facades, which are often weakly visible or invisible in top-down reference maps. They are therefore suitable for evaluating whether DECO can alleviate the co-visibility gap by suppressing geometrically non-co-visible keypoints in challenging oblique-view urban scenes. The 2.5D reference data of AnyVisLoc are reconstructed from aerial photogrammetry, providing accurate geospatial reference data for evaluation. Since AnyVisLoc is originally designed for a retrieval-and-matching localization pipeline, its reference maps usually cover a much larger area than the UAV field of view. Directly matching the UAV image against the full reference map would couple matching with a large search range, making it difficult to analyze the matching performance independently. To avoid this interference, we use the ground-truth pose to rotate the reference map to the same yaw angle as the UAV image, and then crop a reference patch whose field of view is approximately 1.5 times that of the UAV image. The centers of the cropped reference patch and the UAV image are nearly aligned in geographic space. This setting preserves moderate spatial uncertainty while keeping the evaluation focused on cross-view matching.

For OrthoLoC, we follow the original benchmark design and evaluate the dataset under two settings, namely \emph{same-domain} and \emph{cross-domain}, with representative examples shown in Figure~\ref{fig:dataset_overview}(e) and Figure~\ref{fig:dataset_overview}(f), respectively. In the \emph{same-domain} setting, the 2.5D reference data are also generated from aerial photogrammetry. As a result, the UAV images and the reference data are relatively consistent in modality and acquisition time, leading to similar appearance. In contrast, the \emph{cross-domain} setting uses reference data from external sources, such as public geoportals, and therefore introduces stronger modality and temporal discrepancies. This setting is useful for evaluating the robustness of DECO under domain shifts. Since OrthoLoC has already adjusted the overlapping field of view between the UAV image and the reference map to a comparable range, we directly use the provided image pairs without additional reference-map cropping. In addition, based on the ground-truth yaw angles provided by OrthoLoC, we constrain the yaw difference between the UAV image and the reference map to be within $45^\circ$. This step reduces the influence of large in-plane rotation, to which many image matching methods are sensitive, and allows the evaluation to focus more on the effectiveness of the proposed method. Most OrthoLoC samples are close to nadir views, with pitch angles mainly between $60^\circ$ and $90^\circ$. Compared with AnyVisLoc, OrthoLoC provides a larger number of samples and includes more suburban and natural scenes. These characteristics  make it suitable for evaluating the generalization ability of visual localization methods under diverse low-altitude conditions.

\begin{table}[htb]
    \caption{Summary of the datasets used in our experiments.}
    \centering
    \begin{threeparttable}
    \resizebox{\textwidth}{!}{
    \begin{tabular}{c c c c c c c}
        \toprule
        Dataset & Split & \makecell{Reference map \\ resolution (m/px)} & Image number & Altitude range (m) & \makecell{Pitch distribution \\ (0--30$^\circ$ / 30--60$^\circ$ / 60--90$^\circ$)} & Focal length (pixel) \\
        \midrule
        \multirow{4}{*}{AnyVisLoc}
        & scene1      & 0.030 & 300  & 144--303 & 101 / 149 / 50  & 3703/8081/19312 \\
        & scene2  & 0.045 & 256  & 87--214  & 234  / 22 / 0   & 3703/8084/19341 \\
        & scene3     & 0.054 & 252  & 92--120  & 0 / 47 / 205    & 2316             \\
        & scene4       & 0.084 & 1684 & 116--363 & 1 / 1194 / 489  & 3700/3940        \\
        \midrule
        \multirow{2}{*}{OrthoLoC}
        & same-domain  & 0.041--0.731 & 10,923 & 23--201 & 2 / 1156 / 9765 & 566--937 \\
        & cross-domain & 0.041--0.731 & 5,504  & 23--201 & 0 / 705 / 4799 & 566--937 \\
        \bottomrule
    \end{tabular}}
    \end{threeparttable}
    \label{tab:dataset_info}
\end{table}


\subsubsection{Evaluation Metrics}

We evaluate the proposed method from two complementary aspects: geolocalization accuracy and the geometric consistency of matched correspondences. Since this work focuses on UAV planar localization rather than attitude estimation, the localization error is measured by the horizontal position error. Let $\hat{\mathbf{c}}_i=[\hat{x}_i,\hat{y}_i]^{\top}$ and $\mathbf{c}_i=[x_i,y_i]^{\top}$ denote the estimated and ground-truth planar positions of the $i$-th test sample, respectively. The horizontal localization error is defined as $e_i = \left\| \hat{\mathbf{c}}_i - \mathbf{c}_i \right\|_2.$ Based on this error, the localization success rate under a threshold $T$ is defined as
\begin{equation}
\mathrm{SR}(T) = \frac{1}{N}\sum_{i=1}^{N} \mathbb{I}(e_i \leq T),
\label{eq:success_rate}
\end{equation}
where $N$ is the total number of test samples and $\mathbb{I}(\cdot)$ is the indicator function. In the following tables, we report $\mathrm{SR}(1\,\mathrm{m})$ and $\mathrm{SR}(3\,\mathrm{m})$ as T@1 and T@3, respectively. Note that unsuccessful localization cases are also included in the denominator, so this metric does not overestimate performance by counting only successfully localized samples.

In addition to localization success rate, we further use the inlier ratio in the PnP stage to evaluate the geometric consistency of the matched correspondences. Let $m_i$ denote the number of correspondences used for PnP in the $i$-th sample, and let $n_i$ denote the number of inliers identified by RANSAC. The inlier ratio of this sample is defined as $r_i = n_i / m_i$. The average inlier ratio over the test set is then computed as
\begin{equation}
\bar{r} = \frac{1}{N_s}\sum_{i=1}^{N_s} r_i,
\label{eq:avg_inlier_ratio}
\end{equation}
where $N_s$ is the number of samples that successfully enter the PnP stage. A higher inlier ratio indicates better geometric consistency of the 2D--3D correspondences and directly reflects the effectiveness of the proposed co-visibility-aware keypoint selection in suppressing geometrically non-co-visible correspondences.

\subsubsection{Implementation Details}

All experiments were conducted on Ubuntu 22.04 with Python 3.10 and PyTorch. The hardware platform consisted of an NVIDIA GeForce RTX 4090 GPU and an Intel Xeon Platinum 8370C CPU at 2.80 GHz. Unless otherwise stated, the key hyperparameters are set as follows. In the co-visibility-aware matching stage, the geometric co-visibility score and the rank-normalized detector saliency score are coupled according to Equation~\eqref{eq:weighted_keypoint_score}, with $\lambda=1.0$. The number of retained keypoints is set to $K=1024$. In implementation, detector-based methods need to extract a candidate keypoint set much larger than $K$ for subsequent ranking and filtering, which is set to $4K$ in our experiments. In the pose estimation stage, we use \texttt{solvePnPRansac} in OpenCV with a P3P solver and RANSAC. The reprojection error threshold is set to 4 pixels, the maximum number of RANSAC iterations is 2000, and the confidence is set to 0.999.

\subsection{Generality Analysis}
\label{sec:generality}

To evaluate the plug-and-play generality of DECO, we test it with different MDE models, keypoint detectors, and feature matchers. The MDE module is instantiated with five recent models, including Depth Anything 3, UniDepthV2, Depth Pro, MoGe2, and MoGe2-Aerial~\citep{DA3, piccinelli2025unidepthv2, depthpro, moge2, moge2aerial}. For feature extraction, we use SuperPoint, ALIKED, XFeat, and DISK~\citep{detone2018superpoint, zhao2023aliked, potje2024xfeat, DISK}. For feature matching, we evaluate Mutual Nearest Neighbor (MNN), LightGlue, and SuperGlue~\citep{lightglue}. All methods are implemented using official public code and open-source pretrained weights. For brevity, we denote SuperPoint, LightGlue, SuperGlue, and Depth Anything 3 as SP, LG, SG, and DA3, respectively.
\begin{table*}[t]
\centering
\caption{Comparison of different detector--matcher combinations with fixed MDE model DA3.  IR: Inlier Ratio;  Metrics are reported in \%. The \protect\colorbox{bestgreen}{\strut best} results are highlighted.}
\label{tab:anyvisloc_ortholoc_main}
\resizebox{\textwidth}{!}{
\begin{tabular}{lcccccccccccccccccc}
\toprule
\multirow{2}{*}{Method}
& \multicolumn{12}{c}{AnyVisLoc Dataset}
& \multicolumn{6}{c}{OrthoLoC Dataset} \\
\cmidrule(lr){2-13} \cmidrule(lr){14-19}
& \multicolumn{3}{c}{Scene1}
& \multicolumn{3}{c}{Scene2}
& \multicolumn{3}{c}{Scene3}
& \multicolumn{3}{c}{Scene4}
& \multicolumn{3}{c}{Cross-domain}
& \multicolumn{3}{c}{Same-domain} \\
\cmidrule(lr){2-4} \cmidrule(lr){5-7} \cmidrule(lr){8-10} \cmidrule(lr){11-13} \cmidrule(lr){14-16} \cmidrule(lr){17-19}
& T@1 & T@3 & IR
& T@1 & T@3 & IR
& T@1 & T@3 & IR
& T@1 & T@3 & IR
& T@1 & T@3 & IR
& T@1 & T@3 & IR \\
\midrule

SP+MNN
& 39.7 & 72.7 & 12.8
& 7.8 & 19.1 & 2.7
& 11.9 & 50.8 & 6.7
& 22.9 & 69.6 & 8.0
& 8.1 & 20.5 & 5.6
& 45.5 & 55.6 & 24.4 \\
SP+MNN+DECO
& \cellcolor{bestgreen}56.0 & \cellcolor{bestgreen}89.3 & \cellcolor{bestgreen}15.9
& \cellcolor{bestgreen}11.7 & \cellcolor{bestgreen}28.5 & \cellcolor{bestgreen}3.8
& \cellcolor{bestgreen}16.7 & \cellcolor{bestgreen}63.1 & \cellcolor{bestgreen}7.7
& \cellcolor{bestgreen}26.1 & \cellcolor{bestgreen}73.9 & \cellcolor{bestgreen}8.8
& \cellcolor{bestgreen}10.5 & \cellcolor{bestgreen}24.9 & \cellcolor{bestgreen}6.4
& \cellcolor{bestgreen}47.5 & \cellcolor{bestgreen}57.2 & \cellcolor{bestgreen}27.0 \\
\midrule

SP+LG
& 71.7 & 95.3 & 22.6
& 24.6 & 48.4 & 20.0
& 23.0 & 77.8 & 14.1
& \cellcolor{bestgreen}44.6 & 90.0 & 12.2
& 30.7 & 52.4 & 33.1
& 55.1 & 64.1 & 49.9 \\
SP+LG+DECO
& \cellcolor{bestgreen}79.3 & \cellcolor{bestgreen}99.7 & \cellcolor{bestgreen}25.4
& \cellcolor{bestgreen}35.5 & \cellcolor{bestgreen}58.6 & \cellcolor{bestgreen}23.6
& \cellcolor{bestgreen}25.4 & \cellcolor{bestgreen}80.6 & \cellcolor{bestgreen}14.7
& 44.3 & \cellcolor{bestgreen}91.8 & \cellcolor{bestgreen}13.4
& \cellcolor{bestgreen}32.6 & \cellcolor{bestgreen}53.0 & \cellcolor{bestgreen}36.6
& \cellcolor{bestgreen}55.7 & \cellcolor{bestgreen}64.5 & \cellcolor{bestgreen}53.7 \\
\midrule

SP+SG
& 77.0 & 97.3 & 22.2
& 22.3 & 44.9 & 24.3
& 23.8 & 78.2 & 13.6
& 38.5 & 89.1 & 10.9
& 30.9 & 53.3 & 39.4
& 57.6 & 67.1 & 56.4 \\
SP+SG+DECO
& \cellcolor{bestgreen}82.3 & \cellcolor{bestgreen}99.0 & \cellcolor{bestgreen}23.8
& \cellcolor{bestgreen}29.7 & \cellcolor{bestgreen}57.8 & \cellcolor{bestgreen}24.5
& \cellcolor{bestgreen}30.6 & \cellcolor{bestgreen}84.1 & \cellcolor{bestgreen}14.4
& \cellcolor{bestgreen}43.4 & \cellcolor{bestgreen}91.3 & \cellcolor{bestgreen}12.5
& \cellcolor{bestgreen}32.4 & \cellcolor{bestgreen}54.3 & \cellcolor{bestgreen}42.9
& \cellcolor{bestgreen}58.3 & \cellcolor{bestgreen}67.9 & \cellcolor{bestgreen}60.7 \\
\midrule

ALIKED+MNN
& 37.7 & 62.0 & 11.1
& 12.9 & 30.5 & 4.2
& 17.5 & 53.2 & 7.4
& 51.1 & 87.8 & 13.4
& 10.4 & 28.8 & 6.9
& 51.5 & 62.1 & 31.6 \\
ALIKED+MNN+DECO
& \cellcolor{bestgreen}67.7 & \cellcolor{bestgreen}94.0 & \cellcolor{bestgreen}17.4
& \cellcolor{bestgreen}25.8 & \cellcolor{bestgreen}44.5 & \cellcolor{bestgreen}7.5
& \cellcolor{bestgreen}29.8 & \cellcolor{bestgreen}69.8 & \cellcolor{bestgreen}9.4
& \cellcolor{bestgreen}58.2 & \cellcolor{bestgreen}91.6 & \cellcolor{bestgreen}16.4
& \cellcolor{bestgreen}14.9 & \cellcolor{bestgreen}35.0 & \cellcolor{bestgreen}8.4
& \cellcolor{bestgreen}53.1 & \cellcolor{bestgreen}63.7 & \cellcolor{bestgreen}35.3 \\
\midrule

ALIKED+LG
& 49.0 & 75.7 & 18.6
& 27.3 & 52.3 & 21.1
& 20.2 & 61.1 & 8.2
& 38.1 & 84.3 & 10.4
& 9.2 & 29.2 & 23.9
& 43.5 & 60.5 & 36.9 \\
ALIKED+LG+DECO
& \cellcolor{bestgreen}80.3 & \cellcolor{bestgreen}95.7 & \cellcolor{bestgreen}27.8
& \cellcolor{bestgreen}37.1 & \cellcolor{bestgreen}61.7 & \cellcolor{bestgreen}24.4
& \cellcolor{bestgreen}37.7 & \cellcolor{bestgreen}78.2 & \cellcolor{bestgreen}11.8
& \cellcolor{bestgreen}58.8 & \cellcolor{bestgreen}91.8 & \cellcolor{bestgreen}15.7
& \cellcolor{bestgreen}18.0 & \cellcolor{bestgreen}37.3 & \cellcolor{bestgreen}27.8
& \cellcolor{bestgreen}48.2 & \cellcolor{bestgreen}62.4 & \cellcolor{bestgreen}43.5 \\
\midrule

XFeat+MNN
& 16.7 & 38.7 & 9.1
& 1.2 & 4.3 & 5.3
& 4.0 & 25.0 & 5.7
& 7.9 & 34.6 & 7.3
& 2.9 & 13.0 & 9.0
& 30.0 & 49.3 & 22.2 \\
XFeat+MNN+DECO
& \cellcolor{bestgreen}27.7 & \cellcolor{bestgreen}56.0 & \cellcolor{bestgreen}11.2
& \cellcolor{bestgreen}3.5 & \cellcolor{bestgreen}12.9 & \cellcolor{bestgreen}6.0
& \cellcolor{bestgreen}7.9 & \cellcolor{bestgreen}33.3 & \cellcolor{bestgreen}6.3
& \cellcolor{bestgreen}11.3 & \cellcolor{bestgreen}44.1 & \cellcolor{bestgreen}8.1
& \cellcolor{bestgreen}4.2 & \cellcolor{bestgreen}16.2 & \cellcolor{bestgreen}9.6
& \cellcolor{bestgreen}32.6 & \cellcolor{bestgreen}50.5 & \cellcolor{bestgreen}23.8 \\
\midrule

XFeat+LG
& 34.3 & 68.0 & \cellcolor{bestgreen}18.3
& 9.4 & 21.9 & 18.9
& \cellcolor{bestgreen}12.7 & 36.5 & 6.6
& 14.3 & 57.3 & 7.1
& 9.1 & 27.6 & 34.5
& 40.9 & 57.4 & 43.3 \\
XFeat+LG+DECO
& \cellcolor{bestgreen}51.0 & \cellcolor{bestgreen}87.0 & 16.4
& \cellcolor{bestgreen}11.7 & \cellcolor{bestgreen}32.8 & \cellcolor{bestgreen}19.4
& 10.3 & \cellcolor{bestgreen}50.0 & \cellcolor{bestgreen}7.3
& \cellcolor{bestgreen}18.8 & \cellcolor{bestgreen}66.3 & \cellcolor{bestgreen}8.0
& \cellcolor{bestgreen}10.7 & \cellcolor{bestgreen}29.8 & \cellcolor{bestgreen}36.0
& \cellcolor{bestgreen}42.9 & \cellcolor{bestgreen}58.3 & \cellcolor{bestgreen}46.5 \\
\midrule

DISK+MNN
& 58.0 & 74.7 & 21.1
& 9.8 & 25.0 & 9.9
& 21.4 & 62.7 & 13.0
& 54.6 & 88.5 & 20.2
& 18.9 & 32.2 & 14.9
& 40.8 & 48.6 & 31.7 \\
DISK+MNN+DECO
& \cellcolor{bestgreen}75.7 & \cellcolor{bestgreen}93.7 & \cellcolor{bestgreen}27.2
& \cellcolor{bestgreen}14.8 & \cellcolor{bestgreen}35.2 & \cellcolor{bestgreen}12.2
& \cellcolor{bestgreen}25.0 & \cellcolor{bestgreen}75.0 & \cellcolor{bestgreen}14.6
& \cellcolor{bestgreen}58.9 & \cellcolor{bestgreen}91.3 & \cellcolor{bestgreen}22.6
& \cellcolor{bestgreen}21.0 & \cellcolor{bestgreen}35.1 & \cellcolor{bestgreen}17.5
& \cellcolor{bestgreen}41.7 & \cellcolor{bestgreen}49.7 & \cellcolor{bestgreen}34.2 \\
\midrule

DISK+LG
& 64.7 & 87.7 & 25.7
& 24.2 & 48.0 & 27.4
& 25.4 & 75.0 & 13.4
& 54.0 & 88.4 & 16.7
& 17.9 & 33.8 & 33.4
& 47.4 & 57.2 & 54.3 \\
DISK+LG+DECO
& \cellcolor{bestgreen}78.3 & \cellcolor{bestgreen}94.0 & \cellcolor{bestgreen}29.9
& \cellcolor{bestgreen}28.1 & \cellcolor{bestgreen}55.9 & \cellcolor{bestgreen}29.9
& \cellcolor{bestgreen}30.6 & \cellcolor{bestgreen}80.2 & \cellcolor{bestgreen}14.8
& \cellcolor{bestgreen}58.9 & \cellcolor{bestgreen}92.3 & \cellcolor{bestgreen}18.8
& \cellcolor{bestgreen}22.4 & \cellcolor{bestgreen}38.1 & \cellcolor{bestgreen}36.4
& \cellcolor{bestgreen}49.0 & \cellcolor{bestgreen}58.6 & \cellcolor{bestgreen}57.7 \\
\bottomrule
\end{tabular}
}
\end{table*}
We first fix the MDE model as DA3 and compare different detector--matcher combinations. As reported in Table~\ref{tab:anyvisloc_ortholoc_main}, adding DECO improves almost all pipelines on both AnyVisLoc and OrthoLoC. With SP+LG, for example, DECO increases T@3 by 4.4\%, 10.2\%, 2.8\%, and 1.8\% on the four AnyVisLoc scenes, and by 0.6\% and 0.4\% on the OrthoLoC cross-domain and same-domain settings, respectively. Similar trends can be observed for other detector--matcher combinations. The inlier ratio also improves in most cases, indicating that DECO improves the geometric consistency of matched 2D--3D correspondences by filtering out geometrically non-co-visible keypoints before matching. In addition, the improvement is more pronounced on AnyVisLoc than on OrthoLoC. This is mainly because AnyVisLoc contains more oblique UAV images and richer vertical structures, which create a stronger co-visibility gap. Under this condition, DECO can more effectively improve cross-view matching by retaining keypoints that are both visually distinctive and geometrically consistent with the reference map. On OrthoLoC, the localization accuracy in the cross-domain setting is lower than that in the same-domain setting due to significant modality differences between UAV images and reference maps. Nevertheless, DECO still brings stable improvements on OrthoLoC, showing its robustness under different viewing conditions and data sources. 

To further evaluate the generality of DECO with respect to the monocular depth estimation model, we fix the image matching pipeline to SP+LG and compare five different MDE models, including Depth Anything 3, UniDepthV2, Depth Pro, MoGe2, and MoGe2-Aerial. The results are summarized in Table~\ref{tab:depth_model_sp_lg}. DECO consistently improves the SP+LG baseline across different depth estimators on most evaluation metrics, demonstrating that the proposed framework is not tied to a specific MDE model. Meanwhile, the performance variations across different depth estimators indicate that the quality and generalization ability of the estimated depth still affect the effectiveness of the geometric prior.

\begin{table*}[t]
\centering
\caption{Ablation study of different monocular depth estimation models with a fixed SP+LG image matching pipeline. IR: Inlier Ratio; metrics are reported in \%. The \protect\colorbox{bestgreen}{\strut best}, \protect\colorbox{secondgreen}{\strut second-best}, and \protect\colorbox{thirdyellow}{\strut third} results are highlighted.}
\label{tab:depth_model_sp_lg}
\resizebox{\textwidth}{!}{
\begin{tabular}{lcccccccccccccccccc}
\toprule
\multirow{2}{*}{Method}
& \multicolumn{12}{c}{AnyVisLoc Dataset}
& \multicolumn{6}{c}{OrthoLoC Dataset} \\
\cmidrule(lr){2-13} \cmidrule(lr){14-19}
& \multicolumn{3}{c}{Scene1}
& \multicolumn{3}{c}{Scene2}
& \multicolumn{3}{c}{Scene3}
& \multicolumn{3}{c}{Scene4}
& \multicolumn{3}{c}{Cross-domain}
& \multicolumn{3}{c}{Same-domain} \\
\cmidrule(lr){2-4} \cmidrule(lr){5-7} \cmidrule(lr){8-10}
\cmidrule(lr){11-13} \cmidrule(lr){14-16} \cmidrule(lr){17-19}
& T@1 & T@3 & IR
& T@1 & T@3 & IR
& T@1 & T@3 & IR
& T@1 & T@3 & IR
& T@1 & T@3 & IR
& T@1 & T@3 & IR \\
\midrule

SP+LG
& 71.7 & 95.3 & 22.6
& 24.6 & 48.4 & 20.0
& 23.0 & 77.8 & 14.1
& \cellcolor{secondgreen}44.6 & 90.0 & 12.2
& 30.7 & 52.4 & 33.1
& 55.1 & 64.1 & 49.9 \\

SP+LG+DA3+DECO
& \cellcolor{thirdyellow}79.3
& \cellcolor{bestgreen}99.7
& \cellcolor{thirdyellow}25.4
& \cellcolor{secondgreen}35.5
& \cellcolor{bestgreen}58.6
& \cellcolor{bestgreen}23.6
& 25.4
& 80.6
& \cellcolor{thirdyellow}14.7
& \cellcolor{thirdyellow}44.3
& \cellcolor{secondgreen}91.8
& \cellcolor{thirdyellow}13.4
& \cellcolor{secondgreen}32.6
& \cellcolor{secondgreen}53.0
& \cellcolor{secondgreen}36.6
& 55.7
& \cellcolor{thirdyellow}64.5
& 53.7 \\

SP+LG+UniDepthV2+DECO
& \cellcolor{bestgreen}80.0
& \cellcolor{secondgreen}99.0
& 24.6
& 27.0
& 52.7
& 21.2
& \cellcolor{secondgreen}28.6
& \cellcolor{bestgreen}83.3
& 14.6
& 43.4
& \cellcolor{thirdyellow}91.7
& 13.2
& 31.4
& 52.3
& \cellcolor{thirdyellow}36.0
& \cellcolor{bestgreen}56.0
& \cellcolor{bestgreen}64.7
& 53.4 \\

SP+LG+DepthPro+DECO
& \cellcolor{secondgreen}79.7
& \cellcolor{thirdyellow}98.3
& \cellcolor{bestgreen}25.6
& \cellcolor{thirdyellow}34.0
& \cellcolor{thirdyellow}57.4
& \cellcolor{thirdyellow}23.0
& \cellcolor{bestgreen}29.0
& \cellcolor{secondgreen}82.9
& \cellcolor{bestgreen}15.2
& 43.3
& \cellcolor{bestgreen}92.2
& \cellcolor{secondgreen}13.6
& 31.7
& 52.2
& \cellcolor{bestgreen}37.0
& \cellcolor{thirdyellow}55.8
& 64.2
& \cellcolor{bestgreen}54.2 \\

SP+LG+MoGe2+DECO
& 76.1
& 96.6
& \cellcolor{thirdyellow}25.4
& \cellcolor{thirdyellow}34.2
& \cellcolor{secondgreen}58.2
& \cellcolor{secondgreen}23.1
& 23.7
& 80.2
& \cellcolor{thirdyellow}14.7
& 44.2
& 90.6
& \cellcolor{bestgreen}13.7
& \cellcolor{thirdyellow}32.1
& \cellcolor{secondgreen}53.0
& \cellcolor{bestgreen}37.0
& 55.7
& \cellcolor{secondgreen}64.6
& \cellcolor{secondgreen}54.1 \\

SP+LG+MoGe2-Aerial+DECO
& 75.7
& 96.6
& \cellcolor{secondgreen}25.5
& \cellcolor{bestgreen}37.4
& \cellcolor{thirdyellow}57.8
& \cellcolor{thirdyellow}23.5
& \cellcolor{thirdyellow}27.3
& \cellcolor{thirdyellow}81.3
& \cellcolor{secondgreen}14.9
& \cellcolor{bestgreen}46.8
& 89.8
& \cellcolor{secondgreen}13.6
& \cellcolor{bestgreen}33.2
& \cellcolor{bestgreen}53.1
& \cellcolor{bestgreen}37.0
& \cellcolor{secondgreen}55.9
& \cellcolor{secondgreen}64.6
& \cellcolor{thirdyellow}53.9 \\

\bottomrule
\end{tabular}
}
\end{table*}

\begin{figure*}[t]
    \centering
    \includegraphics[width=\textwidth]{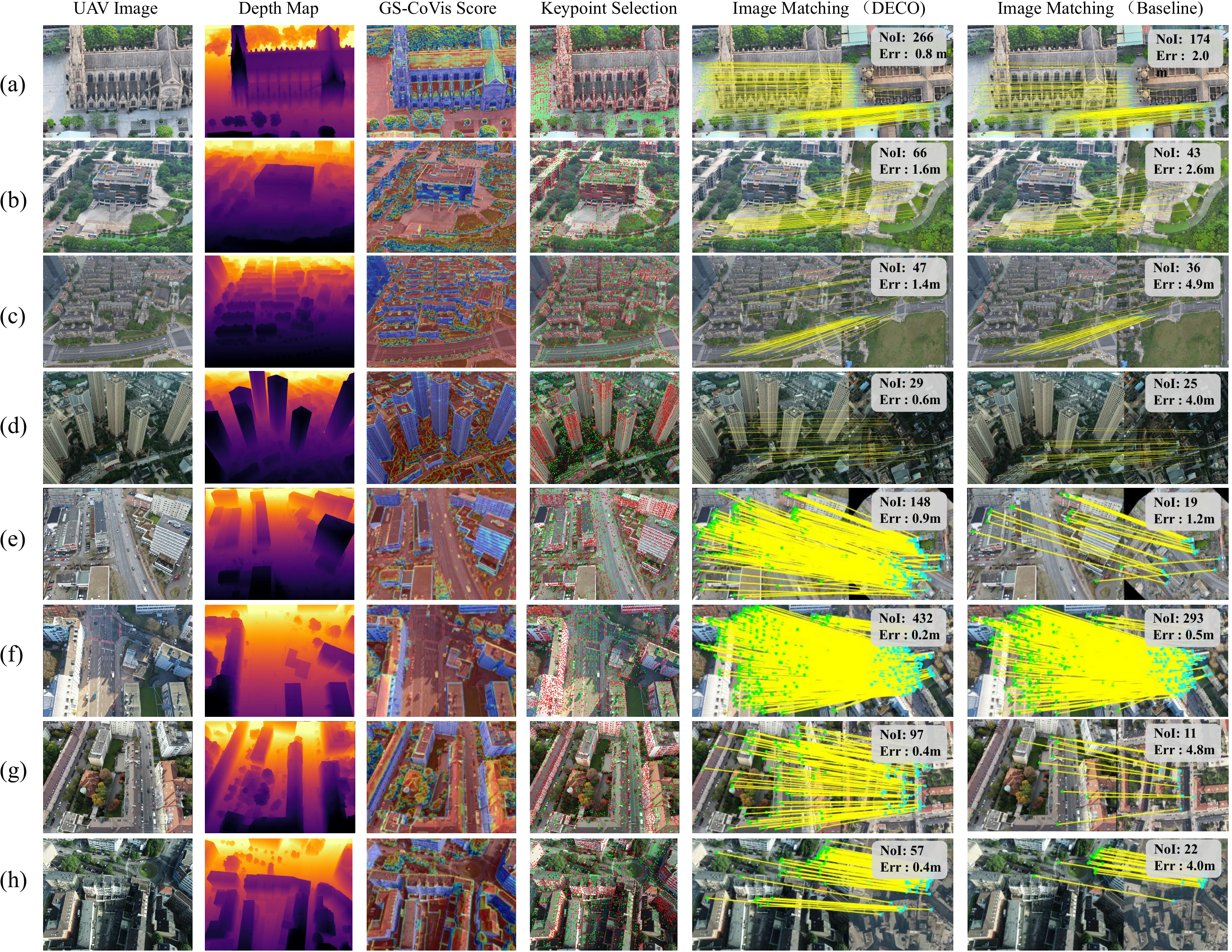}
    \caption{Qualitative visualization of DECO. (a)--(d): AnyVisLoc Scene1--Scene4; (e) and (f): OrthoLoC same-domain examples; (g) and (h): OrthoLoC cross-domain examples. For each row, from left to right: UAV image, monocular depth map, GS-CoVis score heatmap, keypoints retained by DECO, and final matching/localization results of the baseline and DECO. Warmer colors in the heatmap indicate higher GS-CoVis scores. Green points indicate retained keypoints, while red points indicate filtered keypoints. NoI denotes the number of PnP inliers, and Err denotes the localization error.}
    \label{fig:vis_results}
\end{figure*}

To further illustrate the effect of DECO, Figure~\ref{fig:vis_results}(a-h) shows representative results from AnyVisLoc and OrthoLoC under the DA3+SP+LG setting. As shown, SuperPoint tends to detect many keypoints on visually salient but geometrically non-co-visible structures, such as facades, which may introduce mismatches and unstable PnP-RANSAC estimates. In contrast, DECO ranks UAV keypoints using monocular-depth-derived surface geometry and retains those with higher GS-CoVis scores. The resulting correspondences are cleaner and more geometrically consistent, leading to more PnP inliers and more accurate localization.

\subsection{Comparison with State-of-the-Art Methods}
To benchmark DECO against prior-guided matching strategies, we compare it with representative methods that introduce additional priors before or during image matching, including the segmentation-based methods MESA \citep{MESA} and D-MESA \citep{dmesa}, as well as the monocular-depth-guided methods SCE \citep{wang2023guiding} and LiftFeat \citep{LiftFeat}. MESA and D-MESA improve correspondence estimation by introducing region-level semantic priors with the Segment Anything Model (SAM) \citep{kirillov2023segment}, while SCE and LiftFeat exploit monocular-depth-derived geometric cues in the matching process. Since these methods are closely related to our motivation of reducing cross-view ambiguity beyond pure image matching, they provide meaningful baselines for comparison. Following the generalization analysis, we use SP+LG+DA3 as the default setting of DECO. We also report SP+SG+DA3+DECO to provide a fair comparison with existing methods under the same detector-based matching pipeline.

Table~\ref{tab:sota_comparison} reports the comparison results. Overall, DECO achieves highly competitive performance across both AnyVisLoc and OrthoLoC. Although the segmentation-based methods MESA and D-MESA generally improve the accuracy of SP+SG by introducing semantic region priors, their overall performance is still inferior to DECO. This suggests that semantic priors can help reduce some ambiguous matches, but are still limited under extreme viewpoint changes, where semantically corresponding regions may be difficult to align reliably across UAV images and orthographic reference maps. In contrast, DECO reasons about geometric co-visibility before feature matching and ranks keypoints according to monocular-depth-guided surface geometry, making it better suited to the cross-view localization setting. A similar observation can be made for monocular-depth-guided methods. Although LoFTR+SCE introduces surface-curvature cues derived from monocular depth to guide local feature matching, its performance is not consistently better than the original LoFTR baseline. For instance, it slightly improves Scene3 and Scene4, but degrades on Scene1 and Scene2. This indicates that injecting geometric cues into the matching stage alone may not be sufficient, because many erroneous matches originate from features detected on geometrically non-co-visible regions. DECO addresses this issue more directly by filtering these regions before matching. LiftFeat also exploits monocular-depth-derived geometry, but in a different manner by enhancing local feature representation. However, the standalone LiftFeat+MNN pipeline, although better than some lightweight detectors such as XFeat, does not show a clear advantage over stronger detectors such as SuperPoint or DISK (see Table~\ref{tab:anyvisloc_ortholoc_main}). After combining LiftFeat with DECO, the localization performance is consistently improved, indicating that DECO is complementary to geometry-aware local descriptors.

\begin{table*}[t]
\centering
\caption{Comparison with state-of-the-art methods on AnyVisLoc and OrthoLoC. Metrics are reported in \%. The \protect\colorbox{bestgreen}{\strut best}, \protect\colorbox{secondgreen}{\strut second-best}, and \protect\colorbox{thirdyellow}{\strut third} results are highlighted.}

\label{tab:sota_comparison}
\setlength{\tabcolsep}{3.5pt}
\resizebox{\textwidth}{!}{
\begin{tabular}{lcccccccccccc}
\hline
\multirow{3}{*}{Method}
& \multicolumn{8}{c}{AnyVisLoc Dataset}
& \multicolumn{4}{c}{OrthoLoC Dataset} \\
\cmidrule(lr){2-9} \cmidrule(lr){10-13}
& \multicolumn{2}{c}{Scene1}
& \multicolumn{2}{c}{Scene2}
& \multicolumn{2}{c}{Scene3}
& \multicolumn{2}{c}{Scene4}
& \multicolumn{2}{c}{Cross-domain}
& \multicolumn{2}{c}{Same-domain} \\
\cmidrule(lr){2-3} \cmidrule(lr){4-5} \cmidrule(lr){6-7} \cmidrule(lr){8-9} \cmidrule(lr){10-11} \cmidrule(lr){12-13}
& T@1 & T@3 & T@1 & T@3 & T@1 & T@3 & T@1 & T@3 & T@1 & T@3 & T@1 & T@3 \\
\hline

\multicolumn{13}{c}{\textit{Baseline}} \\
\hline
SP+SG
& 77.0 & 97.3 & 22.3 & 44.9 & 23.8 & 78.2 & 38.5 & 89.1 & 30.9 & 53.3 & 57.6 & \cellcolor{thirdyellow}67.1 \\

LoFTR
& 78.6 & 97.8
& \cellcolor{secondgreen}32.1 & 52.2
& 26.3 & 79.5
& 40.2 & \cellcolor{thirdyellow}91.4
& 31.5 & 53.0
& \cellcolor{secondgreen}59.2 & 66.5 \\

\hline

\multicolumn{13}{c}{\textit{Segmentation-based methods}} \\
\hline
SP+SG+MESA
& 79.2 & 97.9
& 25.7 & 49.8
& \cellcolor{thirdyellow}27.6 & 80.1
& 41.5 & 91.2
& 31.8 & 52.7
& \cellcolor{thirdyellow}58.8 & 66.9 \\

SP+SG+D-MESA
& \cellcolor{secondgreen}80.7 & \cellcolor{thirdyellow}98.4
& 27.9 & \cellcolor{thirdyellow}52.3
& 26.4 & \cellcolor{thirdyellow}81.6
& 40.7 & 90.8
& \cellcolor{thirdyellow}32.2 & \cellcolor{thirdyellow}53.4
& 57.6 & \cellcolor{secondgreen}67.4 \\

\hline

\multicolumn{13}{c}{\textit{Monocular-depth-guided methods}} \\
\hline
LoFTR+SCE
& 76.4 & 96.0
& \cellcolor{thirdyellow}31.3 & 51.6
& \cellcolor{secondgreen}30.1 & \cellcolor{secondgreen}82.4
& \cellcolor{thirdyellow}41.8 & \cellcolor{bestgreen}91.9
& 31.2 & \cellcolor{secondgreen}53.6
& \cellcolor{bestgreen}59.4 & 66.8 \\

LiftFeat+MNN
& 28.3 & 48.0 & 10.2 & 19.1 & 19.8 & 53.2 & 20.8 & 69.1 & 4.8 & 17.4 & 33.5 & 49.5 \\
\hline

\multicolumn{13}{c}{\textit{Our methods}} \\
\hline
LiftFeat+MNN+DECO
& 50.3 & 84.0 & 8.6 & 30.9 & 17.9 & 61.9 & 31.4 & 77.3 & 7.2 & 21.8 & 36.3 & 50.7 \\

SP+SG+DECO
& \cellcolor{bestgreen}82.3 & \cellcolor{secondgreen}99.0
& 29.7 & \cellcolor{secondgreen}57.8
& \cellcolor{bestgreen}30.6 & \cellcolor{bestgreen}84.1
& \cellcolor{secondgreen}43.4 & 91.3
& \cellcolor{secondgreen}32.4 & \cellcolor{bestgreen}54.3
& 58.3 & \cellcolor{bestgreen}67.9 \\

SP+LG+DECO
& \cellcolor{thirdyellow}79.3 & \cellcolor{bestgreen}99.7
& \cellcolor{bestgreen}35.5 & \cellcolor{bestgreen}58.6
& 25.4 & 80.6
& \cellcolor{bestgreen}44.3 & \cellcolor{secondgreen}91.8
& \cellcolor{bestgreen}32.6 & 53.0
& 55.7 & 64.5 \\
\hline
\end{tabular}}
\end{table*}

\section{Discussion}
\subsection{Sensitivity Analysis}
\label{sec:sensiti}

\begin{figure*}[h]
    \centering
    \subfigure[Sensitivity to $K$.]{
        \includegraphics[width=0.48\linewidth]{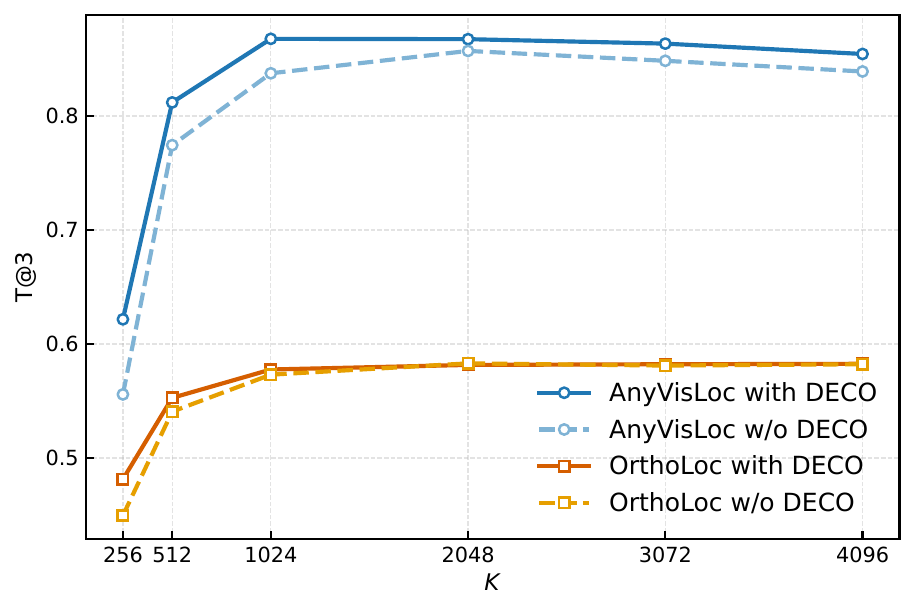}
        \label{fig:sensitivity_k}
    }
    \hfill
    \subfigure[Sensitivity to $\lambda$.]{
        \includegraphics[width=0.48\linewidth]{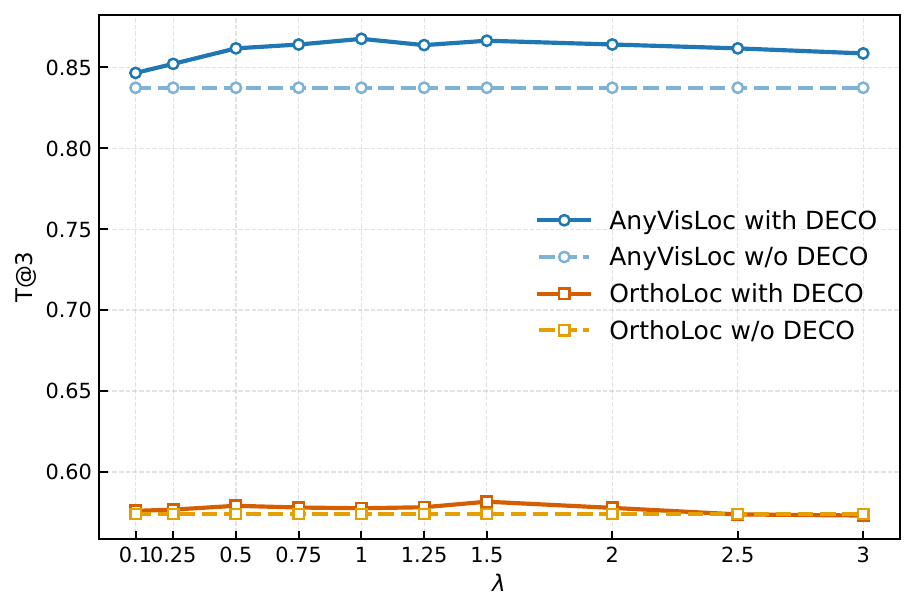}
        \label{fig:sensitivity_lambda}
    }
    \caption{Sensitivity analysis of the score-guided keypoint selection in DECO with respect to the maximum number of retained keypoints $K$ and the geometric weighting exponent $\lambda$ in the proposed GS-CoVis score.}
    \label{fig:sensitivity_analysis}
\end{figure*}

This section analyzes the sensitivity of DECO to two hyperparameters under the DA3+SP+LG setting: the maximum number of retained keypoints $K$ and the geometric weighting exponent $\lambda$ in the proposed GS-CoVis score. All other settings are kept fixed. For the analysis of $K$, we set $\lambda=1$ and vary $K$ from 256 to 4096. For the analysis of $\lambda$, we set $K=1024$ and vary $\lambda$ from 0.1 to 3. The results are shown in Figure~\ref{fig:sensitivity_analysis}.

As shown in Figure~\ref{fig:sensitivity_k}, increasing $K$ first substantially improves localization performance and then leads to gradual saturation. Since DECO selects the top-$K$ UAV keypoints according to the GS-CoVis score, $K$ determines how many score-ranked keypoints that are both visually distinctive and geometrically consistent with the reference map are passed to matching. When $K=256$, the retained set is too sparse to provide sufficient effective 2D--3D correspondences, limiting the robustness of matching and PnP-based pose estimation. Increasing $K$ to 512 and 1024 introduces more high-score correspondences and therefore improves geometric verification. Beyond $K=1024$, newly added keypoints are lower in the score ranking and provide limited additional co-visible evidence, while increasing the matching and geometric verification cost. Therefore, the default setting $K=1024$ used in the main experiments provides a reasonable trade-off between localization accuracy and computational efficiency.

Figure~\ref{fig:sensitivity_lambda} further examines the role of $\lambda$ in the proposed GS-CoVis score. In the score formulation, $\lambda$ controls how strongly the geometric co-visibility prior modulates detector saliency. When $\lambda$ is too small, such as 0.1 or 0.25, the ranking becomes close to saliency-dominated keypoint selection. In this case, visually distinctive but geometrically non-co-visible regions, such as building facades, may still be retained, which limits the benefit of DECO in UAV-to-map localization. As $\lambda$ increases to a moderate range, e.g., 0.5--1.5, the GS-CoVis score better balances geometric co-visibility and visual distinctiveness, leading to consistently strong localization performance. However, when $\lambda$ becomes too large, such as $\lambda \geq 2.5$, the ranking may be overly dominated by the geometric prior, suppressing visually distinctive keypoints that are still useful for feature matching. This can reduce descriptor discriminability and weaken pose stability. Overall, these results indicate that neither detector saliency nor geometric visibility alone is sufficient for robust cross-view matching. Instead, explicitly coupling them through the GS-CoVis score enables DECO to retain keypoints that are both visually distinctive and geometrically consistent. The default setting $\lambda=1$ used in the main experiments lies in this stable and effective range.

\subsection{Effect of Reference-Map Curation and Localization Uncertainty}
\label{sec:curation_discussion}

The dataset curation adopted in Section~4.1.1 reduces the influence of large spatial search ranges and orientation differences, allowing the evaluation to focus on the cross-view co-visibility problem addressed by DECO. To examine the effect of these choices, we further evaluate DECO under relaxed spatial and orientation constraints using DA3+SP+LG. For AnyVisLoc, we enlarge the reference-patch scale from $1.5\times$ to $2.5\times$, remove ground-truth yaw alignment, and combine both changes, thereby increasing spatial and orientation uncertainty. For OrthoLoC, we remove the $45^\circ$ yaw-difference constraint and preserve the original orientation difference of each image pair. The results are summarized in Table~\ref{tab:curation_stress_test}.

\begin{table*}[t]
\centering
\caption{Localization results under the original and relaxed reference-map curation settings using DA3+SP+LG. IR: Inlier Ratio; metrics are reported in \%.}
\label{tab:curation_stress_test}
\resizebox{\textwidth}{!}{
\begin{tabular}{llcccccccccccccccccc}
\toprule
\multirow{2}{*}{Setting}
& \multirow{2}{*}{Method}
& \multicolumn{12}{c}{AnyVisLoc Dataset}
& \multicolumn{6}{c}{OrthoLoC Dataset} \\
\cmidrule(lr){3-14} \cmidrule(lr){15-20}
&
& \multicolumn{3}{c}{Scene1}
& \multicolumn{3}{c}{Scene2}
& \multicolumn{3}{c}{Scene3}
& \multicolumn{3}{c}{Scene4}
& \multicolumn{3}{c}{Cross-domain}
& \multicolumn{3}{c}{Same-domain} \\
\cmidrule(lr){3-5} \cmidrule(lr){6-8} \cmidrule(lr){9-11} \cmidrule(lr){12-14}
\cmidrule(lr){15-17} \cmidrule(lr){18-20}
&
& T@1 & T@3 & IR
& T@1 & T@3 & IR
& T@1 & T@3 & IR
& T@1 & T@3 & IR
& T@1 & T@3 & IR
& T@1 & T@3 & IR \\
\midrule

\multirow{2}{*}{Main curation}
& SP+LG
& 71.7 & 95.3 & 22.6
& 24.6 & 48.4 & 20.0
& 23.0 & 77.8 & 14.1
& 44.6 & 90.0 & 12.2
& 30.7 & 52.4 & 33.1
& 55.1 & 64.1 & 49.9 \\

& SP+LG+DECO
& \textbf{79.3} & \textbf{99.7} & \textbf{25.4}
& \textbf{35.5} & \textbf{58.6} & \textbf{23.6}
& \textbf{25.4} & \textbf{80.6} & \textbf{14.7}
& 44.3 & \textbf{91.8} & \textbf{13.4}
& \textbf{32.6} & \textbf{53.0} & \textbf{36.6}
& \textbf{55.7} & \textbf{64.5} & \textbf{53.7} \\
\midrule

\multirow{2}{*}{$2.5\times$, yaw aligned}
& SP+LG
& 29.3 & 67.0 & 11.9
& 16.8 & 40.2 & 15.4
& 12.7 & 50.0 & 5.7
& \textbf{16.9} & 64.1 & 6.5
& -- & -- & --
& -- & -- & -- \\

& SP+LG+DECO
& \textbf{53.0} & \textbf{85.3} & \textbf{12.6}
& \textbf{26.2} & \textbf{59.8} & \textbf{16.7}
& \textbf{17.5} & \textbf{69.1} & \textbf{6.5}
& 16.7 & \textbf{66.3} & \textbf{7.1}
& -- & -- & --
& -- & -- & -- \\
\midrule

\multirow{2}{*}{$1.5\times$, yaw unaligned}
& SP+LG
& 27.3 & 32.0 & 13.6
& 3.5 & 7.8 & \textbf{16.2}
& 7.5 & 19.8 & \textbf{12.0}
& 13.7 & 33.5 & 11.7
& -- & -- & --
& -- & -- & -- \\

& SP+LG+DECO
& \textbf{34.0} & \textbf{37.3} & \textbf{14.6}
& \textbf{4.3} & \textbf{8.6} & 15.7
& \textbf{7.9} & \textbf{21.8} & 11.3
& \textbf{15.3} & \textbf{34.2} & \textbf{12.5}
& -- & -- & --
& -- & -- & -- \\
\midrule

\multirow{2}{*}{$2.5\times$, yaw unaligned}
& SP+LG
& 13.0 & 27.7 & \textbf{12.6}
& 1.2 & 5.1 & 14.6
& \textbf{6.0} & 14.3 & 9.6
& 3.6 & 17.2 & 8.9
& -- & -- & --
& -- & -- & -- \\

& SP+LG+DECO
& \textbf{21.7} & \textbf{35.0} & 11.7
& \textbf{4.7} & \textbf{11.7} & \textbf{14.9}
& 5.6 & \textbf{15.5} & \textbf{10.3}
& \textbf{4.4} & \textbf{17.8} & \textbf{9.0}
& -- & -- & --
& -- & -- & -- \\
\midrule

\multirow{2}{*}{Original yaw}
& SP+LG
& -- & -- & --
& -- & -- & --
& -- & -- & --
& -- & -- & --
& 10.9 & 20.1 & 25.2
& 22.9 & 28.3 & 33.4 \\

& SP+LG+DECO
& -- & -- & --
& -- & -- & --
& -- & -- & --
& -- & -- & --
& \textbf{11.3} & \textbf{20.2} & \textbf{27.0}
& \textbf{23.1} & \textbf{28.4} & \textbf{35.4} \\

\bottomrule
\end{tabular}
}
\end{table*}

Relaxing the reference-map curation leads to a clear decrease in absolute localization performance. On AnyVisLoc, both enlarging the search region and removing yaw alignment make localization more difficult, with the strongest degradation when both factors are present. Similarly, removing the yaw-difference constraint on OrthoLoC substantially reduces localization accuracy. These results confirm that the curation adopted in the main experiments effectively reduces interference from search-range and rotation-related factors, enabling a more controlled evaluation of the co-visibility problem.

Despite the increased difficulty, DECO generally retains its advantage over the corresponding baseline across the relaxed settings. The improvements are more pronounced when the search area is enlarged while yaw alignment is retained, and become smaller under large orientation uncertainty, where the rotation robustness of the underlying feature pipeline becomes increasingly important. This indicates that DECO remains effective under increased localization uncertainty, while large orientation differences introduce additional challenges beyond the co-visibility issue targeted by DECO.

\subsection{Discussion on the Gravity-Direction Prior}

As described in Section~\ref{sec:method}, the normal-gravity co-visibility score in Equation~\eqref{eq:normal_gravity_score} relies on the gravity direction $\mathbf{g}_c$. In the main DECO pipeline, $\mathbf{g}_c$ is obtained from the coarse roll and pitch angles according to Equation~\eqref{eq:gravity}. Such attitude priors are usually available from the onboard IMU, making this a practical strategy for gravity direction estimation. However, they may be inaccurate or unavailable when inertial navigation drift is severe or when the timestamps are not synchronized. To handle this setting, we further design a dominant-plane gravity estimation strategy that infers the gravity direction from monocular-depth geometry.

This strategy is motivated by the Manhattan-world assumption commonly observed in urban and suburban environments~\citep{guo2022neural}. In such scenes, large-scale dominant planes often correspond to horizontal support surfaces, such as roads, ground regions, or flat rooftops, whose normals provide a useful cue for the gravity direction. Since the camera-coordinate point cloud $\mathcal{P}_c$ has already been reconstructed in Equation~\eqref{eq:backprojection}, we fit a dominant plane $\pi:\mathbf{n}_{\pi}^{\top}\mathbf{X}+d=0$ in $\mathcal{P}_c$ using RANSAC:
\begin{equation}
(\mathbf{n}_{\pi}^{*},d^{*})
=
\arg\max_{\mathbf{n}_{\pi},d}
\sum_{\mathbf{X}_i\in\mathcal{P}_c}
\mathbf{1}\!\left(
\frac{|\mathbf{n}_{\pi}^{\top}\mathbf{X}_i+d|}{\|\mathbf{n}_{\pi}\|_2}
< \epsilon
\right),
\label{eq:gravity_plane_ransac}
\end{equation}
where $\epsilon$ denotes the plane inlier threshold. The camera-frame gravity direction is then approximated by the normalized dominant-plane normal:
\begin{equation}
\hat{\mathbf{g}}_c =
\frac{\mathbf{n}_{\pi}^{*}}{\|\mathbf{n}_{\pi}^{*}\|_2}.
\label{eq:plane_gravity}
\end{equation}

\begin{figure}[t]
\centering
\includegraphics[width=0.7\linewidth]{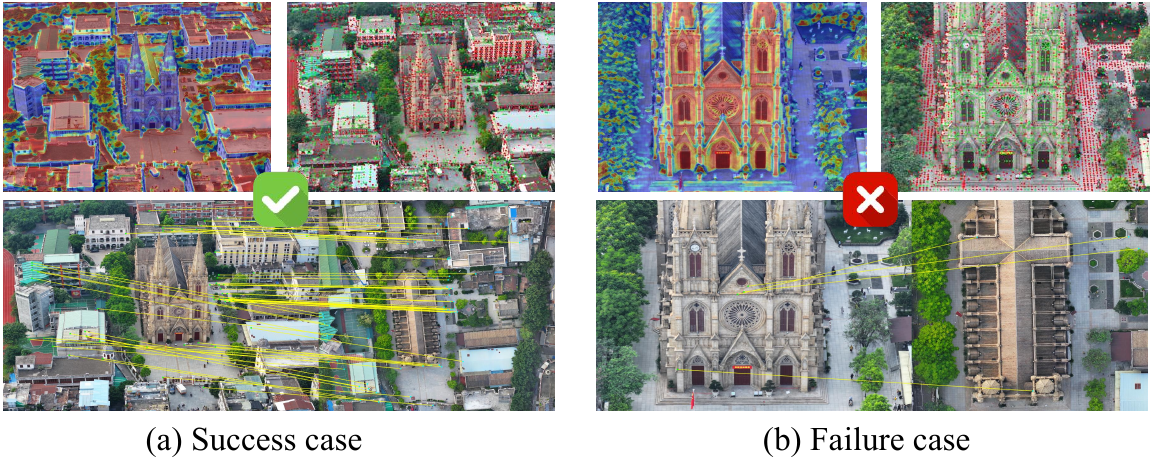}
\caption{Representative success and failure cases when the gravity direction is estimated using the dominant-plane instead of using the attitude prior. Warmer colors in the heatmap indicate higher GS-CoVis scores. Green points indicate retained keypoints, while red points indicate filtered keypoints.}
\label{fig:plane_normal_cases}
\end{figure}

This dominant-plane estimation method only changes how the gravity direction is obtained, while other steps in Section~\ref{sec:method} remain unchanged. Table~\ref{tab:gravity_estimation} compares the localization performance obtained with different gravity-direction estimation strategies. Compared with the SP+LG baseline, the plane-normal strategy improves nearly all metrics, demonstrating that monocular-depth geometry can provide an effective substitute when attitude priors are unavailable. Nevertheless, the attitude-prior variant remains stronger overall. This is expected because dominant-plane estimation depends on the quality of monocular depth and the scene layout. When the UAV image contains a clear ground or roof-like support surface, the fitted plane normal is well aligned with the true gravity direction. However, when the image is dominated by vertical facades or dense vegetation, the dominant fitted plane may deviate from the horizontal support surface. Representative cases in Figure~\ref{fig:plane_normal_cases} further illustrate this behavior. Overall, dominant-plane gravity estimation provides a practical fallback for DECO when reliable attitude priors are unavailable.

\begin{table*}[h]
\centering
\caption{Comparison of localization performance obtained with different gravity-direction estimation strategies. IR: Inlier Ratio. The \protect\colorbox{bestgreen}{\strut best}, \protect\colorbox{secondgreen}{\strut second-best}, and \protect\colorbox{thirdyellow}{\strut third} results are highlighted.}
\label{tab:gravity_estimation}
\resizebox{\textwidth}{!}{
\begin{tabular}{lcccccccccccccccccc}
\toprule
\multirow{2}{*}{Method}
& \multicolumn{12}{c}{AnyVisLoc Dataset}
& \multicolumn{6}{c}{OrthoLoC Dataset} \\
\cmidrule(lr){2-13} \cmidrule(lr){14-19}
& \multicolumn{3}{c}{Scene1}
& \multicolumn{3}{c}{Scene2}
& \multicolumn{3}{c}{Scene3}
& \multicolumn{3}{c}{Scene4}
& \multicolumn{3}{c}{Cross-domain}
& \multicolumn{3}{c}{Same-domain} \\
\cmidrule(lr){2-4} \cmidrule(lr){5-7} \cmidrule(lr){8-10} \cmidrule(lr){11-13} \cmidrule(lr){14-16} \cmidrule(lr){17-19}
& T@1 & T@3 & IR & T@1 & T@3 & IR & T@1 & T@3 & IR & T@1 & T@3 & IR & T@1 & T@3 & IR & T@1 & T@3 & IR \\
\midrule
SP+LG (Baseline) 
& \cellcolor{thirdyellow}71.7 & \cellcolor{thirdyellow}95.3 & \cellcolor{thirdyellow}22.6 
& \cellcolor{thirdyellow}24.6 & \cellcolor{thirdyellow}48.4 & \cellcolor{thirdyellow}20.0 
& \cellcolor{thirdyellow}23.0 & \cellcolor{thirdyellow}77.8 & \cellcolor{thirdyellow}14.1 
& \cellcolor{secondgreen}44.6 & \cellcolor{secondgreen}90.0 & \cellcolor{thirdyellow}12.2 
& \cellcolor{thirdyellow}30.7 & \cellcolor{thirdyellow}52.4 & \cellcolor{thirdyellow}33.1 
& \cellcolor{thirdyellow}55.1 & \cellcolor{thirdyellow}64.1 & \cellcolor{thirdyellow}49.9 \\

SP+LG+DA3+DECO (Attitude prior)
& \cellcolor{secondgreen}79.3 & \cellcolor{bestgreen}99.7 & \cellcolor{bestgreen}25.4 
& \cellcolor{bestgreen}35.5 & \cellcolor{bestgreen}58.6 & \cellcolor{bestgreen}23.6 
& \cellcolor{bestgreen}25.4 & \cellcolor{secondgreen}80.6 & \cellcolor{bestgreen}14.7 
& \cellcolor{thirdyellow}44.3 & \cellcolor{bestgreen}91.8 & \cellcolor{secondgreen}13.4 
& \cellcolor{bestgreen}32.6 & \cellcolor{bestgreen}53.0 & \cellcolor{bestgreen}36.6 
& \cellcolor{bestgreen}55.7 & \cellcolor{bestgreen}64.5 & \cellcolor{bestgreen}53.7 \\

SP+LG+DA3+DECO (Plane-normal estimation) 
& \cellcolor{bestgreen}81.7 & \cellcolor{secondgreen}96.3 & \cellcolor{secondgreen}25.3 
& \cellcolor{secondgreen}29.7 & \cellcolor{secondgreen}57.0 & \cellcolor{secondgreen}22.2 
& \cellcolor{secondgreen}25.3 & \cellcolor{bestgreen}81.0 & \cellcolor{secondgreen}14.6 
& \cellcolor{bestgreen}44.9 & \cellcolor{thirdyellow}89.2 & \cellcolor{bestgreen}13.5 
& \cellcolor{secondgreen}32.1 & \cellcolor{secondgreen}52.7 & \cellcolor{secondgreen}36.4 
& \cellcolor{secondgreen}55.6 & \cellcolor{secondgreen}64.3 & \cellcolor{secondgreen}53.5 \\

\bottomrule
\end{tabular}
}
\end{table*}

\section{Conclusion}

This paper presented DECO, a depth-guided co-visibility reasoning framework for low-altitude UAV visual localization. The key insight of DECO is that  reliable cross-view UAV-to-map matching should jointly consider detector saliency and geometric co-visibility. By explicitly coupling these two factors, DECO shifts keypoint selection from appearance-driven detection to geometry-aware correspondence construction, suppressing visually salient yet geometrically non-co-visible structures before matching. Extensive experiments demonstrate that DECO consistently improves localization accuracy across different MDE models, feature detectors, and feature matchers. Compared with recent prior-guided matching methods, DECO provides a plug-and-play solution that more directly addresses the co-visibility gap in low-altitude UAV-to-map localization.

Despite these advantages, DECO still has several limitations. Its performance remains affected by the accuracy and generalizability of the adopted MDE model on UAV imagery. In addition, explicitly introducing an MDE module increases the computational cost of the visual localization pipeline. Future work will focus on learning feature detectors under the supervision of co-visible region masks, so that geometrically co-visible features can be extracted more directly without explicitly introducing an MDE module. Another promising direction is to explore tighter coupling of depth and semantic priors for more robust co-visible region reasoning under complex low-altitude scenes.

\bibliographystyle{tfcad}
\bibliography{ref}

\end{document}